\documentclass[letterpaper, 10 pt, conference]{ieeeconf}
\IEEEoverridecommandlockouts                              
\usepackage{times}
\usepackage{epsfig}
\usepackage{graphicx}
\usepackage{amsmath}
\usepackage{amssymb}
\usepackage{multirow}

\usepackage{graphicx}
\usepackage{epsfig}
\usepackage{epic,eepic}
\usepackage{times}
\usepackage{mathptmx}
\usepackage{amsfonts}
\usepackage{amsmath}
\usepackage{cite}
\usepackage{color}

\usepackage{times}
\usepackage{multicol}

\definecolor{red}{rgb}{1,0,0}
\definecolor{green}{rgb}{0,1,0}
\definecolor{blue}{rgb}{0,0,1}
\definecolor{violet}{rgb}{1,0,1}
\definecolor{cyan}{cmyk}{1,0,0,0}
\definecolor{magenta}{cmyk}{0,1,0,0}
\definecolor{yellow}{cmyk}{0,0,1,0}

\definecolor{white}{rgb}{1,1,1}

\usepackage{jumoline}

\newcommand{\CO}[1]{}

\newcommand{\CommentOut}[1]{}

 \newcommand{\editage}[1]{}

\begin{document}

\newcommand{\FIG}[3]{
\begin{minipage}[b]{#1cm}
\begin{center}
\includegraphics[width=#1cm]{#2}\\
{\scriptsize #3}
\end{center}
\end{minipage}
}

\newcommand{\FIGU}[3]{
\begin{minipage}[b]{#1cm}
\begin{center}
\includegraphics[width=#1cm,angle=180]{#2}\\
{\scriptsize #3}
\end{center}
\end{minipage}
}

\newcommand{\FIGm}[3]{
\begin{minipage}[b]{#1cm}
\begin{center}
\includegraphics[width=#1cm]{#2}\\
{\scriptsize #3}
\end{center}
\end{minipage}
}

\newcommand{\FIGR}[3]{
\begin{minipage}[b]{#1cm}
\begin{center}
\includegraphics[angle=-90,width=#1cm]{#2}
\\
{\scriptsize #3}
\vspace*{1mm}
\end{center}
\end{minipage}
}

\newcommand{\FIGRpng}[5]{
\begin{minipage}[b]{#1cm}
\begin{center}
\includegraphics[bb=0 0 #4 #5, angle=-90,clip,width=#1cm]{#2}\vspace*{1mm}
\\
{\scriptsize #3}
\vspace*{1mm}
\end{center}
\end{minipage}
}

\newcommand{\FIGpng}[5]{
\begin{minipage}[b]{#1cm}
\begin{center}
\includegraphics[bb=0 0 #4 #5, clip, width=#1cm]{#2}\vspace*{-1mm}\\
{\scriptsize #3}
\vspace*{1mm}
\end{center}
\end{minipage}
}

\newcommand{\FIGtpng}[5]{
\begin{minipage}[t]{#1cm}
\begin{center}
\includegraphics[bb=0 0 #4 #5, clip,width=#1cm]{#2}\vspace*{1mm}
\\
{\scriptsize #3}
\vspace*{1mm}
\end{center}
\end{minipage}
}

\newcommand{\FIGRt}[3]{
\begin{minipage}[t]{#1cm}
\begin{center}
\includegraphics[angle=-90,clip,width=#1cm]{#2}\vspace*{1mm}
\\
{\scriptsize #3}
\vspace*{1mm}
\end{center}
\end{minipage}
}

\newcommand{\FIGRm}[3]{
\begin{minipage}[b]{#1cm}
\begin{center}
\includegraphics[angle=-90,clip,width=#1cm]{#2}\vspace*{0mm}
\\
{\scriptsize #3}
\vspace*{1mm}
\end{center}
\end{minipage}
}

\newcommand{\FIGC}[5]{
\begin{minipage}[b]{#1cm}
\begin{center}
\includegraphics[width=#2cm,height=#3cm]{#4}~$\Longrightarrow$\vspace*{0mm}
\\
{\scriptsize #5}
\vspace*{8mm}
\end{center}
\end{minipage}
}

\newcommand{\FIGf}[3]{
\begin{minipage}[b]{#1cm}
\begin{center}
\fbox{\includegraphics[width=#1cm]{#2}}\vspace*{0.5mm}\\
{\scriptsize #3}
\end{center}
\end{minipage}
}

\newcommand{\acprPaperID}{25}

\title{\LARGE \bf 
Dark Reciprocal-Rank: Boosting Graph-Convolutional Self-Localization Network via Teacher-to-student Knowledge Transfer 
}

\author{%
~~~ Takeda Koji ~~~~~ Tanaka Kanji
\thanks{Our work has been supported in part by 
JSPS KAKENHI 
Grant-in-Aid 
for Scientific Research (C) 17K00361, and (C) 20K12008.}
\thanks{%
K. Takeda is with Graduate School of Engineering, University of Fukui, Japan. 
K. Tanaka is with Faculty of Engineering, University of Fukui, Japan. 
{\tt\small \{takedakoji00, tanakakanji\}@gmail.com}
}
}

\maketitle{}

\begin{abstract}
In visual robot self-localization, graph-based scene representation and matching have recently attracted research interest as robust and discriminative methods for self-localization. Although effective, their computational and storage costs do not scale well to large-size environments. To alleviate this problem, we formulate self-localization as a graph classification problem and attempt to use the graph convolutional neural network (GCN) as a graph classification engine. A straightforward approach is to use visual feature descriptors that are employed by state-of-the-art self-localization systems, directly as graph node features. However, their superior performance in the original self-localization system may not necessarily be replicated in GCN-based self-localization. To address this issue, we introduce a novel teacher-to-student knowledge-transfer scheme based on rank matching, in which the reciprocal-rank vector output by an off-the-shelf state-of-the-art teacher self-localization model is used as the dark knowledge to transfer. Experiments indicate that the proposed graph-convolutional self-localization network can significantly outperform state-of-the-art self-localization systems, as well as the teacher classifier. 
\end{abstract}

\section{Introduction}

In visual robot self-localization, graph-based scene representation and matching have attracted recent research interest as robust and discriminative methods for self-localization. For example, in \cite{seqgraph}, a multi-view self-localization application was addressed by representing each view image frame as a graph node and by connecting neighboring image frames via  graph edges. In \cite{x-view}, a single-view self-localization application was addressed by representing semantically segmented regions as graph nodes and connecting neighboring segments via graph edges. In these applications, a query scene graph is matched against each map graph according to the similarity of graph node descriptors (e.g., image descriptors \cite{Arandjelovic16}, %
region descriptors \cite{convnet15}) 
and the graph structure. Although they are effective, their computational and storage costs increase in proportion to the environment size and do not scale well to large environments.

\newcommand{\figA}{
  \begin{figure}[t]
\centering
\FIG{8.5}{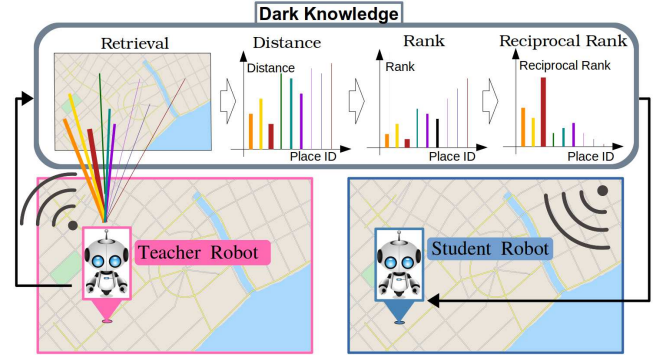}{}\vspace*{-3mm}\\
\caption{We propose the use of the reciprocal-rank vector as the dark knowledge to be transferred from a self-localization model (i.e., teacher) to a graph convolutional self-localization network (i.e., student), for improving the self-localization performance.}\label{fig:A}
\vspace*{-3mm}
\end{figure}
}

\figA

To alleviate this problem, we formulate self-localization as a graph classification task and use the graph convolutional neural network (GCN) as a graph classification engine. Our approach, wherein the GCN is used as a scene graph classifier, is analogous to the recent paradigm of using a convolutional neural network (CNN) as a scene image classifier \cite{itsc19}. We inherit desirable properties of the classification task formulation, such as the flexibility in defining place classes \cite{pp}, 
compressed classifier model \cite{compactkd}, 
and high classification speed \cite{speedupkd}. 
A key difference from the image classifier tasks is that the input visual data must be translated to graph data before being input to the GCN. This problem is the main focus of the present study.

A straightforward approach is to employ visual feature descriptors used by state-of-the-art self-localization systems, such as CNN-based \cite{Arandjelovic16}, 
GAN-based \cite{feat2}, 
and autoencoder-based features \cite{feat1}, 
directly as graph node features. However, the main concern is that visual feature descriptors are not optimized for graph convolutions. In theory, their superior performance in the original self-localization system may not necessarily be replicated in GCN-based self-localization. Our experimental results indicated that the self-localization performance deteriorated when visual feature descriptors were directly used as a graph node feature descriptor in the GCN model.

To address this issue, we introduce a novel teacher-to-student knowledge-transfer scheme based on rank matching \cite{darkrank} (Fig. \ref{fig:A}). 
The basic idea is to introduce a state-of-the-art self-localization model (e.g., bag-of-words image retrieval \cite{ibowlcd}, 
object proposal and matching \cite{convnet15}, 
and deep image feature descriptor \cite{Arandjelovic16}) 
as a teacher classifier. This approach is inspired by the rank-matching loss used by recent transfer-learning schemes \cite{darkrank}, where rank values are employed as the dark knowledge transferred from the teacher classifier to the student classifier. While rank transfer has been used in transfer learning with CNNs, its use in feature transfer with GCNs is non-trivial and was addressed for the first time in this study.

The main contributions of this work are summarized as follows: (1) We propose a novel graph node descriptor, which transfers the prediction of an off-the-shelf state-of-the-art teacher self-localization model to the student GCN classifier. (2) We show that a class-specific reciprocal-rank vector is a proper and effective representation of the dark knowledge to transfer. (3) We experimentally show that the proposed graph-convolutional self-localization network can significantly outperform state-of-the-art self-localization systems, as well as the teacher classifier.

\section{Scene Graph Model}

In the proposed GCN self-localization framework, two types of scene graph models are used: the single-view subimage-level scene graph (SVSL) and the multi-view image-level scene graph (MVIL), which are described in Sections \ref{sec:svsl} and \ref{sec:mvil}, respectively.

\newcommand{\figC}{
\begin{figure}[t]
\centering
\FIG{8.5}{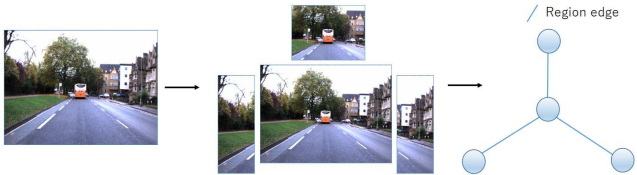}{}\vspace*{-3mm}\\
\caption{Single-view subimage-level scene graph (SVSL).}\label{fig:C}
\vspace*{-3mm}
\end{figure}
}

\figC

\subsection{Single-view Subimage-level Scene Graph (SVSL)}\label{sec:svsl}

The SVSL takes as input a single-view scene and converts it into a subimage-level scene graph. In the implementation, we use a total of four subimage nodes that correspond to the entire image region [0, 0]$\times$[1080, 800] and three bounding boxes: CENTER [270, 200]$\times$[810, 600], RIGHT [780, 0]$\times$[1080, 800], and LEFT [0, 0]$\times$[300, 800]. As shown in Fig. \ref{fig:C}, in an SVSL graph, the edges extend in a star shape from the entire image node to the other three subimage nodes. While this scene graph model requires only a single-view image as an input, the invariance of the graph depends significantly on the invariance of the image segmentation. This limitation does not affect the MVIL model (\ref{sec:mvil}).

It can be effective to use a semantic segmentation (SS) technique to decompose an image into subimages instead of fixed bounding boxes. For example, in \cite{x-view}, the method of decomposing a scene into subimages via SS and connecting the segmented subimages via object-level edges experimentally worked well under an ideal condition of ground-truth segmentations. However, the good performance was not replicated in our current implementation of GCN self-localization. In a preliminary experiment, we attempted to use a state-of-the-art SS technique \cite{ourss} instead of the fixed segmentation strategy, which was significantly affected by segmentation noise.

\subsection{Multi-view Image-level Scene Graph (MVIL)}\label{sec:mvil}

The MVIL takes as input a view-sequence scene and represents it as an image-level scene graph with multi-attribute frame image nodes and two types of graph edges: time and attribute edges (Fig. \ref{fig:D}). As an example, in experiments, we consider at most $K=4$ different attribute images, which are obtained by converting an original input RGB image with $K$ different image filters: Canny, depth regression, and SS, as shown in Fig. \ref{fig:D}. A time edge connects the nodes of successive image frames with the same attribute. An attribute edge connects different attribute nodes of the same image frame. $(K-1)$ attribute images are connected to the RGB image node via attribute edges, yielding a star shape from the RGB image node to the $(K-1)$ attribute image nodes, as shown in Fig. \ref{fig:D}. To facilitate the invariance of the time edge, the sampling of image frames is controlled so that the travel distance between successive image frames (connected by a time edge) becomes constant with regard to odometry measurements. While a multi-view scene graph requires as input a view-sequence, it is largely unaffected by segmentation noise. This is an appealing property of the MVIL model.

\newcommand{\figD}{
\begin{figure}[t]
\centering
\FIG{8.5}{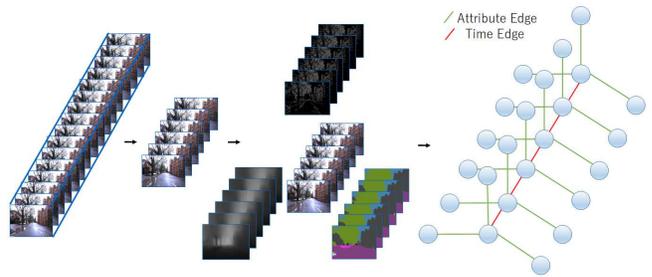}{}\vspace*{-3mm}\\
\caption{Multi-view image-level scene graph (MVIL).}\label{fig:D}
\vspace*{-3mm}
\end{figure}
}

\figD

The implementation details are as follows: We implemented $(K-1)$ types of attribute images: Canny, depth, and SS images, which were converted from an original RGB image by using the Canny edge detector \cite{canny}, deep depth regressor \cite{alhashim2018high}, and Deeplab v3+ \cite{deeplab3}, respectively. The weight parameters of the deep depth regressor and Deeplab v3+ were pretrained on the KITTI dataset \cite{kitti} and the Cityscapes dataset \cite{cityscapes}, respectively. The pixel values of an SS image were defined by the color map in \cite{cityscapes}. The length of the view-sequence for the MVIL model was 10 frames, with 2-m intervals.

\section{GCN Self-localization}

In this section, we describe the proposed framework for lightweight and accurate self-localization based on knowledge transfer. In the proposed framework, a lightweight representation of the scene graph is obtained using knowledge transfer from an external teacher self-localization model. Additionally, the accuracy of self-localization can be higher than that of the teacher self-localization model.
 
\subsection{Knowledge Transfer}

In knowledge transfer \cite{distil}, the prediction results of a teacher model are often used as the dark knowledge to transfer. In particular, we propose the use of the (reciprocal-) rank vector as the representation of such dark knowledge. Many off-the-shelf self-localization systems (e.g., bag-of-words systems \cite{ibowlcd}, classification systems \cite{sc}, and map-matching systems \cite{mm}) can be modeled as ranking systems. Therefore, our (reciprocal-) rank-based scheme has a broader application area than existing knowledge-transfer schemes, e.g., those where intermediate signals of the teacher systems are used as the dark knowledge to transfer.

As an example, in our experimental system, a typical nearest neighbor (NN) image classifier with a NetVLAD image descriptor \cite{alhashim2018high} was employed as the teacher model. The teacher model used a visual image as an input, computed the L2 nearest-neighbor distance from the query image descriptor to the class-specific database image descriptors, and then converted the distance values into a class-specific rank value vector; a smaller rank value corresponded to a better degree of matching. Such pairings of the input image and output rank value vector are used as the dark knowledge to transfer in our scheme. We observed that a reciprocal-rank value vector is a good representation of an attribute-image-node descriptor, as discussed in Section \ref{sec:exp}.

\subsection{GCN Classifier}

This subsection describes the procedure for graph convolution, focusing on the equation for forward propagation. A scene graph is represented as $G=(V,E)$, where $V$ represents the set of nodes and $E$ represents the set of edges. Let $v_i\in V$ denote a node and $e_{ij}=(v_i, v_j)\in E$ denote an edge pointing from $v_j$ to $v_i$. The graph is defined as an undirected graph; i.e., whenever $e_{ij}$ exists, $e_{ji}$ exists. The neighborhood of a node $v$ is defined as $N(v)=\{u\in V | (u, v) \in E\}$. Each node $v$ has a feature vector $h\in R^D$, where $D$ is the number of dimensions of the feature vector. We performed an experimental ablation study, in which not only the class-specific reciprocal-rank vector but also the other intermediate representations, such as the original NetVLAD vector, class-specific NN-distance vector, and class-specific rank vector, were considered as the node feature descriptor.

The graph convolution operation takes node $v_i$ in the graph and processes it in the following manner. First, it receives messages from nodes connected by the edge. Then, the collected messages are summed via the SUM function. The result is passed through a single-layer fully connected neural network followed by a nonlinear transformation for conversion into a new feature vector. In this study, we used the rectified linear unit (ReLU) operation as the nonlinear transformation, which is expressed as follows:
\begin{equation}
{\bf h}_{i}^{new} = \mbox{ReLU}
\left({\bf W} 
\left(\sum_{u \in N(v_i) \cup v_i} {\bf h}_{u}
\right)
\right).
\end{equation}
Here, $W$ represents an $R^{D\times F}$ weight matrix, and $D$ and $F$ represent the numbers of dimensions of the node feature vector before and after the linear transformation, respectively. The foregoing process can be generalized to the processing of node features in the $l$-th GCN layer:
\begin{equation}
{\bf h}_{i}^{(l)} = \mbox{ReLU}
\left({\bf W}^{(l-1)} 
\left(\sum_{u \in N(v_i) \cup v_i} {\bf h}_{u}^{(l-1)}
\right)
\right).
\end{equation}
The process was applied to all the nodes in the graph in each iteration, yielding a new graph that had the same shape as the original graph but updated node features. The iterative process was repeated $L$ times, where $L$ represents the ID of the last GCN layer. After the graph node information obtained in this manner were averaged, the probability value vector of the prediction for the graph was obtained by applying the fully connected layer and the softmax function. This averaging operation is called ``Readout." For the probability value vector of the output $p$, the operation is expressed as follows:
\begin{equation}
\label{eq:gcn_output}
 {\bf p} = \mbox{Softmax}\left(FC\left( \frac{1}{|V|} \sum_{u \in V} {\bf h}_{u}^{L}
\right)\right).
\end{equation}
where $h_u$ is a feature of node $u$ after it passes through the last GCN layer. For implementation, we used the Deep Graph Library \cite{wang2019dgl} on the Pytorch backend.

\section{Experiments}\label{sec:exp}

We conducted self-localization experiments to confirm the effectiveness of the proposed method by using the publicly available Oxford Robotcar Dataset \cite{RobotCarDatasetIJRR}.

\newcommand{\tabA}{
\begin{table}[t]
\centering
\caption{Statistics of the dataset.}\label{tab:A}
  \begin{tabular}{|r||r|r|r|r|}\hline
    date  & weather & \#images & detour & roadworks \\ \hline 
    2015-08-28-09-50-22 & sun & 31,855 & $\times$ & $\times$ \\ \hline
    2015-10-30-13-52-14 & overcast & 48,196 & $\times$ & $\times$ \\ \hline
    2015-11-10-10-32-52 & overcast & 29,350 & $\times$ & $\circ$ \\ \hline
    2015-11-12-13-27-51 & clouds & 41,472 & $\circ$ & $\circ$ \\ \hline
    2015-11-13-10-28-08 & overcast, sun & 42,968 & $\times$ & $\times$  \\ \hline
  \end{tabular}
\end{table}

}

\tabA

\subsection{Settings}

The Oxford Robotcar Dataset was obtained by a robotic vehicle-mounted camera when a robot car traveled along the same route in different seasons and with different weather and lighting conditions. Table \ref{tab:A} presents details regarding the dataset used in this study. The onboard camera used was a PointGreyBumblebeeXB3 (BBX3-13S2C-38) trinocular stereo camera (the center camera, 1280$\times$960$\times$3, 16 Hz). To avoid self-reflections and self-occlusions due to the vehicle, we used an image region of 1080$\times$800 pixels   (with 100 pixels from the left and right and 160 pixels from the bottom removed).

\newcommand{\figB}{
\begin{figure}[t]
\centering
\FIG{3}{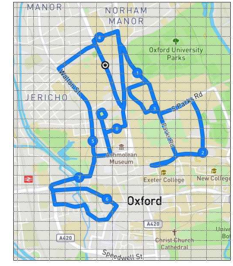}{}\vspace*{-3mm}\\
\caption{Example of place partitioning.}\label{fig:B}
\vspace*{-3mm}
\end{figure}
}

\figB

To define the place class, the workspace of the Oxford Robotcar Dataset was partitioned into a two-dimensional regular grid of place classes according to the ground-truth global positioning system coordinates (Fig. \ref{fig:B}). More formally, an area surrounded by 0.001 degree of latitude and longitude was defined as a one-place class. The number of classes for these test seasons ranged from 82 to 86. The rows indicate the training seasons, and the columns indicate the test seasons. The classes that existed only for training in each season pair and the classes with less than five images in the area were not used in the experiments. ``Unseen" classes, which existed only in the test season, were used as-is. The parameters of the NetVLAD descriptor were trained on the Pittsburgh (Pitts250k) dataset \cite{alhashim2018high}.

For each dataset, all possible overlapping subsequences with travel distance 20 [m] were sampled from the entire image sequence and used as training/test samples. For single-view methods, the first image frame of each subsequence is used as a query input. For multi-view methods, each subsequence is represented by a length 10 view-sequence with 2-m intervals and used as a query input.

The number of GCN layers $L$ was set as 2. The number of dimensions of the intermediate representation was 256. Thus, when the number of classes was $C$, the number of dimensions of the feature vectors (from the bottom layer to the top layer) was $C$$\rightarrow$$256$$\rightarrow$$256$$\rightarrow$$C$. The node aggregation method used the SUM operation and the ReLU activation function. The number of epochs was set as 5. The batch size was 32, and the learning rate was 0.001. The cross-entropy loss function and the Adam optimizer were used.

\subsection{Comparison Methods}

We used NetVLAD \cite{Arandjelovic16} and SeqSLAM \cite{longterm} as comparison methods. The implementation of NetVLAD was based on \cite{cieslewski2018data}.  NetVLAD was used in a single-view image-level self-localization scheme, in which the nearest-neighbor matching of place classes in terms of the Euclidean distance was performed. SeqSLAM was used as a multi-view image-level self-localization scheme. The implementation of SeqSLAM was based on the C++ version of OpenSeqSLAM. The parameters of SeqSLAM were optimized for the Nortland dataset, and no parameter manipulation was performed. The image IDs output by SeqSLAM were converted into the place class IDs to which the images belonged, and then the class IDs were simply used as outputs of the system.

\subsection{Results}

Computation time for GCN classification was 23.8 msec per graph (Intel (R) Xeon (R) GOLC 6130 CPU @ 2.10 GHz). For the GCN training, the speed was satisfactory (170 sec for a size 31,835 training set) even with CPU. This indicates that our approach can be implemented even on low-cost hardware with moderate performance, such as that used by small, inexpensive robots \cite{smallcheap}.

\newcommand{\figI}{
\begin{figure}[t]
\centering
\FIG{8.5}{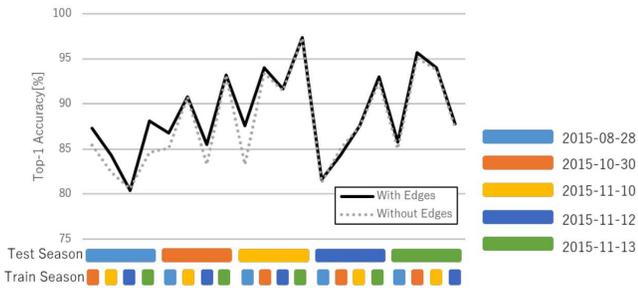}{}\vspace*{-3mm}\\
\caption{Performance results for single-view scene graph with and without edge connections.}\label{fig:I}
\vspace*{-3mm}
\end{figure}
}

\figI

\newcommand{\figH}{
\begin{figure}[t]
\centering
\FIG{8.5}{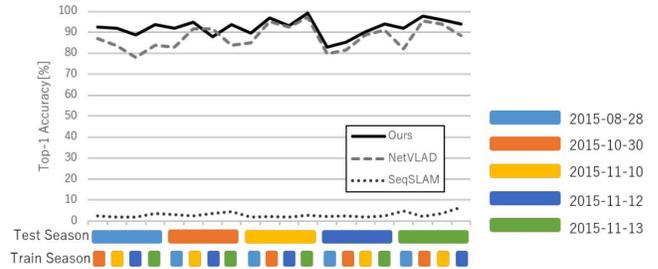}{}\vspace*{-3mm}\\
\caption{Performance results for different training and test season pairs.}\label{fig:H}
\vspace*{-3mm}
\end{figure}
}

\figH

The results for the SVSL scene graph are presented in Fig. \ref{fig:I}. For an ablation study, in addition to the proposed SVSL scene graph, a naive scene graph without edge connections was tested. As shown, the performance was better when edge connections were used.

The results for the proposed and comparison methods are presented in Fig. \ref{fig:H}. The proposed method employs an MVIL scene graph with raw RGB, Canny, and SS attribute image nodes (i.e., $K=3$). First, the result for $K=2$ is shown. In this study, we investigated which node feature is appropriate, which image conversion method is compatible with it, and which graph structure is optimal. Clearly, the proposed method outperforms the comparison methods, i.e., NetVLAD and SeqSLAM.

\section{Conclusions}

We presented a framework for enhancing a visual robot self-localization system using GCN. The proposed framework combines the accuracy of state-of-the-art self-localization systems and the flexibility and efficiency of the GCN. To this end, a novel teacher-to-student knowledge-transfer scheme based on rank matching was introduced. Experimental results indicated that the proposed framework outperformed the state-of-the-art methods and the teacher self-localization system. The reciprocal-rank vector was found to be effective dark knowledge to transfer, and in a future study, we plan to develop additional knowledge-transfer strategies for improving the GCN self-localization performance.

\bibliographystyle{IEEEtran}
\bibliography{gcn20}

\begin{thebibliography}{10}
\providecommand{\url}[1]{#1}
\csname url@rmstyle\endcsname
\providecommand{\newblock}{\relax}
\providecommand{\bibinfo}[2]{#2}
\providecommand\BIBentrySTDinterwordspacing{\spaceskip=0pt\relax}
\providecommand\BIBentryALTinterwordstretchfactor{4}
\providecommand\BIBentryALTinterwordspacing{\spaceskip=\fontdimen2\font plus
\BIBentryALTinterwordstretchfactor\fontdimen3\font minus
  \fontdimen4\font\relax}
\providecommand\BIBforeignlanguage[2]{{%
\expandafter\ifx\csname l@#1\endcsname\relax
\typeout{** WARNING: IEEEtran.bst: No hyphenation pattern has been}%
\typeout{** loaded for the language `#1'. Using the pattern for}%
\typeout{** the default language instead.}%
\else
\language=\csname l@#1\endcsname
\fi
#2}}

\bibitem{seqgraph}
T.~Naseer, L.~Spinello, W.~Burgard, and C.~Stachniss, ``Robust visual robot
  localization across seasons using network flows.'' in \emph{AAAI}, 2014, pp.
  2564--2570.

\bibitem{x-view}
A.~Gawel, C.~Del~Don, R.~Siegwart, J.~Nieto, and C.~Cadena, ``X-view:
  Graph-based semantic multi-view localization,'' \emph{IEEE Robotics and
  Automation Letters}, vol.~3, no.~3, pp. 1687--1694, 2018.

\bibitem{Arandjelovic16}
R.~Arandjelovi\'c, P.~Gronat, A.~Torii, T.~Pajdla, and J.~Sivic, ``{NetVLAD}:
  {CNN} architecture for weakly supervised place recognition,'' in \emph{IEEE
  Conference on Computer Vision and Pattern Recognition}, 2016.

\bibitem{convnet15}
N.~S{\"u}nderhauf, S.~Shirazi, F.~Dayoub, B.~Upcroft, and M.~Milford, ``On the
  performance of convnet features for place recognition,'' in \emph{IEEE/RSJ
  Int. Conf. Intelligent Robots and Systems (IROS)}, 2015, pp. 4297--4304.

\bibitem{itsc19}
N.~Yang, K.~Tanaka, Y.~Fang, X.~Fei, K.~Inagami, and Y.~Ishikawa, ``Long-term
  vehicle localization using compressed visual experiences,'' pp. 2203--2208,
  2018.

\bibitem{pp}
{\"O}.~Erkent and I.~Bozma, ``Place representation in topological maps based on
  bubble space,'' in \emph{2012 IEEE International Conference on Robotics and
  Automation}, 2012, pp. 3497--3502.

\bibitem{compactkd}
T.~Li, J.~Li, Z.~Liu, and C.~Zhang, ``Few sample knowledge distillation for
  efficient network compression,'' in \emph{Proceedings of the IEEE/CVF
  Conference on Computer Vision and Pattern Recognition}, 2020, pp.
  14\,639--14\,647.

\bibitem{speedupkd}
S.~Hofst{\"a}tter, S.~Althammer, M.~Schr{\"o}der, M.~Sertkan, and A.~Hanbury,
  ``Improving efficient neural ranking models with cross-architecture knowledge
  distillation,'' \emph{arXiv preprint arXiv:2010.02666}, 2020.

\bibitem{feat2}
H.~{Hu}, H.~{Wang}, Z.~{Liu}, C.~{Yang}, W.~{Chen}, and L.~{Xie},
  ``Retrieval-based localization based on domain-invariant feature learning
  under changing environments,'' in \emph{IEEE/RSJ Int. Conf. Intelligent
  Robots and Systems (IROS)}, 2019, pp. 3684--3689.

\bibitem{feat1}
N.~Merrill and G.~Huang, ``{CALC2.0}: Combining appearance, semantic and
  geometric information for robust and efficient visual loop closure,'' in
  \emph{IEEE/RSJ Int. Conf. Intelligent Robots and Systems (IROS)}, Macau,
  China, Nov. 2019.

\bibitem{darkrank}
Y.~Chen, N.~Wang, and Z.~Zhang, ``Darkrank: Accelerating deep metric learning
  via cross sample similarities transfer,'' \emph{arXiv preprint
  arXiv:1707.01220}, 2017.

\bibitem{ibowlcd}
E.~Garcia-Fidalgo and A.~Ortiz, ``ibow-lcd: An appearance-based loop-closure
  detection approach using incremental bags of binary words,'' \emph{IEEE
  Robotics and Automation Letters}, vol.~3, no.~4, pp. 3051--3057, 2018.

\bibitem{ourss}
L.-C. Chen, Y.~Zhu, G.~Papandreou, F.~Schroff, and H.~Adam, ``Encoder-decoder
  with atrous separable convolution for semantic image segmentation,'' in
  \emph{Proceedings of the European conference on computer vision (ECCV)},
  2018, pp. 801--818.

\bibitem{canny}
J.~Canny, ``A computational approach to edge detection,'' \emph{IEEE
  Transactions on pattern analysis and machine intelligence}, no.~6, pp.
  679--698, 1986.

\bibitem{alhashim2018high}
I.~Alhashim and P.~Wonka, ``High quality monocular depth estimation via
  transfer learning,'' \emph{arXiv preprint arXiv:1812.11941}, 2018.

\bibitem{deeplab3}
L.-C. Chen, Y.~Zhu, G.~Papandreou, F.~Schroff, and H.~Adam, ``Encoder-decoder
  with atrous separable convolution for semantic image segmentation,'' in
  \emph{Proceedings of the European conference on computer vision (ECCV)},
  2018, pp. 801--818.

\bibitem{kitti}
A.~Geiger, P.~Lenz, and R.~Urtasun, ``Are we ready for autonomous driving? the
  kitti vision benchmark suite,'' in \emph{Conference on Computer Vision and
  Pattern Recognition (CVPR)}, 2012.

\bibitem{cityscapes}
M.~Cordts, M.~Omran, S.~Ramos, T.~Rehfeld, M.~Enzweiler, R.~Benenson,
  U.~Franke, S.~Roth, and B.~Schiele, ``The cityscapes dataset for semantic
  urban scene understanding,'' in \emph{Proceedings of the IEEE conference on
  computer vision and pattern recognition}, 2016, pp. 3213--3223.

\bibitem{distil}
G.~Hinton, O.~Vinyals, and J.~Dean, ``Distilling the knowledge in a neural
  network,'' \emph{arXiv preprint arXiv:1503.02531}, 2015.

\bibitem{sc}
G.~Kim, B.~Park, and A.~Kim, ``1-day learning, 1-year localization: Long-term
  lidar localization using scan context image,'' \emph{IEEE Robotics and
  Automation Letters}, vol.~4, no.~2, pp. 1948--1955, 2019.

\bibitem{mm}
J.~Neira, J.~D. Tard{\'o}s, and J.~A. Castellanos, ``Linear time vehicle
  relocation in slam,'' in \emph{ICRA}.\hskip 1em plus 0.5em minus 0.4em\relax
  Citeseer, 2003, pp. 427--433.

\bibitem{wang2019dgl}
M.~Wang, L.~Yu, D.~Zheng, Q.~Gan, Y.~Gai, Z.~Ye, M.~Li, J.~Zhou, Q.~Huang,
  C.~Ma, Z.~Huang, Q.~Guo, H.~Zhang, H.~Lin, J.~Zhao, J.~Li, A.~J. Smola, and
  Z.~Zhang, ``Deep graph library: Towards efficient and scalable deep learning
  on graphs,'' \emph{ICLR Workshop on Representation Learning on Graphs and
  Manifolds}, 2019.

\bibitem{RobotCarDatasetIJRR}
W.~Maddern, G.~Pascoe, C.~Linegar, and P.~Newman, ``{1 Year, 1000km: The Oxford
  RobotCar Dataset},'' \emph{The International Journal of Robotics Research
  (IJRR)}, vol.~36, no.~1, pp. 3--15, 2017.

\bibitem{longterm}
M.~J. Milford and G.~F. Wyeth, ``Seqslam: Visual route-based navigation for
  sunny summer days and stormy winter nights,'' in \emph{2012 IEEE Int. Conf.
  Robotics and Automation}.\hskip 1em plus 0.5em minus 0.4em\relax IEEE, 2012,
  pp. 1643--1649.

\bibitem{cieslewski2018data}
T.~Cieslewski, S.~Choudhary, and D.~Scaramuzza, ``Data-efficient decentralized
  visual slam,'' in \emph{2018 IEEE International Conference on Robotics and
  Automation (ICRA)}.\hskip 1em plus 0.5em minus 0.4em\relax IEEE, 2018, pp.
  2466--2473.

\bibitem{smallcheap}
H.~Aagela, M.~Al-Nesf, and V.~Holmes, ``An asus\_xtion\_probased indoor mapping
  using a raspberry pi with turtlebot robot turtlebot robot,'' in \emph{2017
  23rd International Conference on Automation and Computing (ICAC)}.\hskip 1em
  plus 0.5em minus 0.4em\relax IEEE, 2017, pp. 1--5.

\end{thebibliography}

\end{document}